\documentclass[letterpaper,journal]{IEEEtran}
\usepackage{amsmath,amsfonts}
\usepackage{algorithm}
\usepackage{amssymb}
\usepackage[spaceRequire=false]{algpseudocodex}

\usepackage{array}
\usepackage{textcomp}
\usepackage{graphicx}
\usepackage{stfloats}
\usepackage{url}

\usepackage{tabularx}
\usepackage{booktabs}
\usepackage{threeparttable}

\usepackage[hidelinks]{hyperref}
\usepackage{cleveref}
\Crefname{figure}{Fig.}{Figs.}

\usepackage{xcolor}
\newcommand{\emp}[1]{\textcolor{blue}{#1}}

\usepackage{xspace}
\makeatletter
\DeclareRobustCommand\onedot{\futurelet\@let@token\@onedot}
\def\@onedot{\ifx\@let@token.\else.\null\fi\xspace}
\def\eg{\emph{e.g}\onedot,~} 
\def\ie{\emph{i.e}\onedot,~} \def\Ie{\emph{I.e}\onedot}
\makeatother

\algnewcommand{\algorithmicgoto}{\textbf{go to}}%
\algnewcommand{\Goto}[1]{\algorithmicgoto~\ref{#1}}%

\usepackage[backend=biber,style=ieee,natbib=true,maxcitenames=2,mincitenames=1,doi=true,isbn=false,url=false,eprint=false]{biblatex}
\begin{document}

\title{Labels Generated by Large Language Models\\Help Measure People's Empathy \textit{in Vitro}}
\author{Md~Rakibul~Hasan,~\IEEEmembership{Graduate~Student~Member,~IEEE,}
Yue Yao,~\IEEEmembership{Graduate~Student~Member,~IEEE,}
Md~Zakir~Hossain,~\IEEEmembership{Member,~IEEE,}
Aneesh~Krishna,
Imre~Rudas,
Shafin~Rahman,
and~Tom~Gedeon,~\IEEEmembership{Senior~Member,~IEEE}
\thanks{M R Hasan, Y Yao, M Z Hossain, A Krishna and T Gedeon are with School of Electrical Engineering, Computing and Mathematical Sciences, Curtin University, Bentley WA 6102, Australia.}
\thanks{I Rudas is with Obuda University, Budapest, Hungary.}
\thanks{S Rahman is with North South University, Dhaka 1229, Bangladesh.}
\thanks{M R Hasan is also with BRAC University, Dhaka 1212, Bangladesh.}
\thanks{Y Yao, M Z Hossain and T Gedeon are also with The Australian National University, Canberra ACT 2600, Australia.}
\thanks{T Gedeon is also with Obuda University, Budapest, Hungary.}
\thanks{E-mail: \{Rakibul.Hasan, Zakir.Hossain1, A.Krishna, Tom.Gedeon\}@curtin.edu.au, yue.yao@anu.edu.au, rudas@uni-obuda.hu, shafin.rahman@northsouth.edu}
\thanks{Corresponding author: M R Hasan}
}

\markboth{IEEE Journal of Selected Topics in Signal Processing}%
{Hasan \MakeLowercase{\textit{et al.}}: Labels Generated by Large Language Models Help Measure People's Empathy \textit{in Vitro}}

\IEEEpubid{This work has been submitted to the IEEE for possible publication. Copyright may be transferred without notice.}
\maketitle

\begin{abstract}
Large language models (LLMs) have revolutionised many fields, with LLM-as-a-service (LLMSaaS) offering accessible, general-purpose solutions without costly task-specific training. In contrast to the widely studied prompt engineering for directly solving tasks (\textit{in vivo}), this paper explores LLMs' potential for \textit{in-vitro} applications: using LLM-generated labels to improve supervised training of mainstream models. We examine two strategies -- (1) noisy label correction and (2) training data augmentation -- in empathy computing, an emerging task to predict psychology-based questionnaire outcomes from inputs like textual narratives. Crowdsourced datasets in this domain often suffer from noisy labels that misrepresent underlying empathy. We show that replacing or supplementing these crowdsourced labels with LLM-generated labels, developed using psychology-based scale-aware prompts, achieves statistically significant accuracy improvements. Notably, the RoBERTa pre-trained language model (PLM) trained with noise-reduced labels yields a state-of-the-art Pearson correlation coefficient of 0.648 on the public NewsEmp benchmarks. This paper further analyses evaluation metric selection and demographic biases to help guide the future development of more equitable empathy computing models. Code and LLM-generated labels are available at \url{https://github.com/hasan-rakibul/LLMPathy}.
\end{abstract}


\section{Introduction}

Large language models (LLMs) have become a go-to approach across a variety of tasks, such as emotion recognition \citep{ziyang2024leveraging} and empathy detection \citep{li2024chinchunmei,kong2024ru}. Due to high computational demands, coupled with environmental impact, training or fine-tuning LLMs often becomes costly. This limitation has led to increasing adoption of LLMs as a service (LLMSaaS), where users access trained LLMs via online APIs with computation on cloud \citep{sun2022black}. LLMSaaS can be utilised \textit{in-vivo}, \ie prompt engineering to directly solve tasks such as named entity recognition \citep{hu2024improving}, sentiment analysis \citep{fei-etal-2023-reasoning} and empathy detection \citep{li2024chinchunmei,kong2024ru}, or \textit{in-vitro} \citep{zheng2017unlabeled}\footnote{Like \citep{zheng2017unlabeled}, we use the term \textit{``in-vitro"} to refer to leveraging LLM outputs out of the box in a different model.}, \ie integrating LLM outputs into \emph{other} models.

\IEEEpubidadjcol 

We are motivated by the following considerations. First, most current applications of LLMSaaS leverage LLM outputs \textit{in-vivo} \citep{hu2024improving,fei-etal-2023-reasoning,li2024chinchunmei,kong2024ru}. We shift to their utility \textit{in-vitro} to fine-tune smaller pre-trained language models (PLMs)\footnote{We use \textit{``pre-trained language models (PLMs)''} to refer specifically to smaller models like the BERT family of models, distinguishing them from LLMs, which are also pre-trained but significantly larger.} like RoBERTa \citep{liu2019roberta}. In particular, we propose to utilise LLMSaaS in a \textit{data-centric AI} approach \citep{mazumder2024dataperf} to \textbf{(1)} enhance the quality of training labels and to \textbf{(2)} increase the amount of quality training data for supervised training of PLMs.

Second, for representation learning, maintaining data quality is critical -- captured succinctly by the phrase, ``garbage in, garbage out'' \citep{geiger2020garbage}. While deep learning research has mostly focused on proposing new algorithms, improvement in data-centric AI is equally important \citep{mazumder2024dataperf}. As a data-centric AI approach, we leverage LLMs to enhance data quality. The effectiveness of our proposed approach is demonstrated in an emerging field -- empathy detection.


Empathy is defined as \textit{``an affective response more appropriate to another's situation than one's own''} \citep{hoffman2000empathy}. In psychology, various questionnaires have been developed to measure empathy. Empathy computing\footnote{We use the terms empathy computing, detection, prediction and measurement interchangeably.}, in computer science, complements these psychology-based methods by aiming to map the questionnaire outcomes from input stimuli such as textual narratives, audiovisual interactions and physiological signals \citep{hasan2024empathy}. One well-known questionnaire is the empathy measurement scale proposed by \citet{batson1987distress}, which assesses empathy across six dimensions: sympathetic, moved, compassionate, tender, warm and soft-hearted.

\begin{figure*}[t!]
    \centering
    \begin{minipage}{0.45\textwidth}
    \centering
    \includegraphics[width=0.9\textwidth]{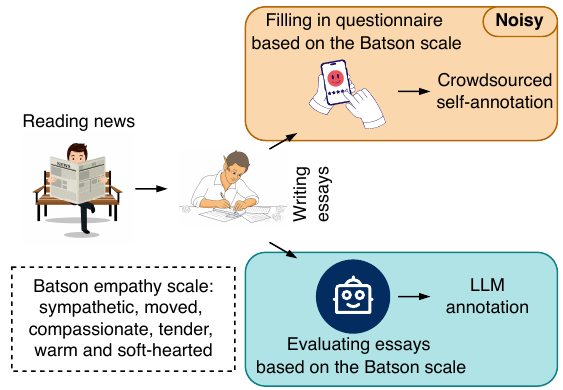}
    \end{minipage}
    \begin{minipage}{0.45\textwidth}
        \centering
        \includegraphics[width=0.9\columnwidth]{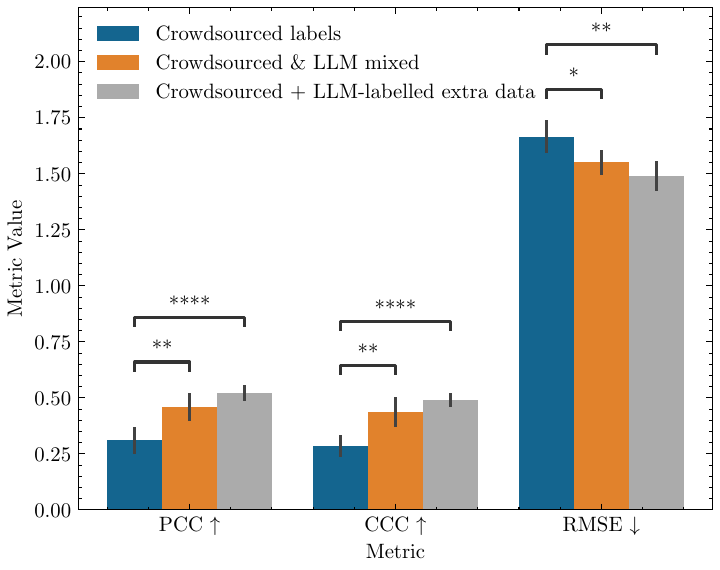}
    \end{minipage}
    \caption{\textbf{Left:} Overview of traditional and LLM-based methods to annotate essays for detecting empathy in the essays written in response to news articles. To be shown in our experiment, existing crowdsourced self-annotation through questionnaires is found to be incorrect in many samples. Our proposed approach involves annotating the essays using LLM, which is then used to reduce label noise and to get additional training data. \textbf{Right:} Impact of our LLM usage is showcased through a performance comparison between existing crowdsourced annotations, LLM-based label noise correction and the inclusion of additional data labelled by LLM. Statistical significance is calculated using \texttt{Statannotations} package \citep{charlier2022statannotations}, where * means $0.01 < p\text{-value} \leq 0.05$, ** means $0.001 < p\text{-value} \leq 0.01$ and **** means $p \text{-value}\leq 0.0001$ (\Ie more * means higher statistical significance).} 
    \label{fig:significance}
\end{figure*}

Empathy \textit{computing} offers the potential to improve people's empathic skills, which in turn strengthens interpersonal relationships across various human interactions \citep{hasan2024empathy}. In healthcare, for example, empathic writing in medical documents (\eg patient reports) can promote understanding and trust between clinicians and patients \citep{jani2012role}. Similarly, in education, written communication like emails and feedback on assignments has become a vital medium for expressing care and addressing students' emotional needs \citep{aldrup2022empathy}. Journalism also demonstrates the importance of empathy in written narratives. For example, a news article on a family’s recovery after a devastating event often goes beyond factual reporting and offers a compassionate perspective that engages readers emotionally and deepens their connection to the news story. Specifically, this paper measures people's empathy in essays written in response to scenarios reported in newspaper articles. 

Empathy is inherently subjective, and machine learning models, including LLMs, used for its detection exhibit biases across different demographic groups \citep{gabriel-etal-2024-ai}. We explore potential sources of such biases (\Cref{subsec:zero-and-bias}). Additionally, while the Pearson correlation coefficient (PCC) remains the most commonly used evaluation metric in empathy computing \citep{hasan2024empathy}, it does not account for the magnitude of the error. To address this, we advocate for adopting the concordance correlation coefficient (CCC) (\Cref{subsec:eval}).

Neural networks are prone to memorising training data, \ie overfitting. This issue is exacerbated by noisy labels, where traditional regularisation techniques like dropout and weight decay often fall short \citep{zhang2021understanding}. A major challenge in ensuring data quality is, therefore, addressing label noise, defined as labels that deviate from their intended values. It is a significant challenge in empathy computing datasets collected through crowdsourcing. Platforms like Amazon Mechanical Turk offer quick access to large participant pools. Accordingly, crowdsourcing with questionnaire-based self-assessment labelling is a popular way of collecting data in computational social psychology and human behaviour research, such as empathy \citep{tafreshi2021wassa} and emotion recognition \citep{mohammad2010emotions}. However, such data often suffer from inaccuracies due to inattentiveness or multitasking among participants, compromising data reliability \citep{sheehan2018crowdsourcing,jia2017using,huang2012detecting,obrochta2021anomalous}. This necessitates strategies to enhance data quality post-collection.



The overarching goal of this paper is to address the question: \textit{``How can LLMs enhance training of PLMs to improve empathy computing accuracy?''} As illustrated in \Cref{fig:significance}, our proposed \textit{in-vitro} applications achieve statistically significant performance improvements. The first application, which automatically adjusts training labels, demonstrates consistent performance improvement across all metrics compared to the baseline PLM trained on the original dataset. The second application, leveraging additional LLM-labelled training data, further enhances model performance, yielding the highest statistically significant performance gains with a $p\text{-value}<0.0001$ \citep{charlier2022statannotations}.

Our key contributions are summarised as follows:
\begin{enumerate}
    \item We propose two \textit{in-vitro} applications of LLMs: mitigating label noise and getting additional training data for PLMs.
    \item We design a novel scale-aware prompt that enables LLMs to annotate data while adhering to annotation protocols grounded in theoretical frameworks.
    \item We investigate challenges in empathy computing datasets and advocate for new evaluation metrics.
    \item Our proposed methods achieve statistically significant performance improvements over the baseline models across multiple datasets and set a new state-of-the-art empathy computing performance.
\end{enumerate}

\section{Related Work}

\subsection{LLM in Data Annotation}
The advent of LLMs has inspired numerous studies exploring LLMs' application in data annotation, often positioning them as a substitute for traditional human annotation. For instance, \citet{niu2024text} examined the potential of LLMs in emotion annotation tasks and reported that LLMs can generate emotion labels closely aligned with human annotations. Similarly, \citet{wang2021want} explored the utility of LLMs in annotating datasets for various natural language processing tasks, including sentiment analysis, question generation and topic classification. While they highlighted the cost-effectiveness of LLM-based annotations, they also noted LLMs' limitations compared to human annotators. Departing from this line of work, our approach explores LLM-generated labels to enhance the training of PLMs. Specifically, we integrate LLM-generated labels with human-generated labels rather than exclusive use of either LLM- or human-generated labels.

We examine our approach in two distinct applications: label adjustment and training data enhancement. Related to our first application, \citet{hasan2024llm-gem} also explored label noise adjustment, but that approach relies on subjects' demographic information (\eg age, gender and race) in the prompting process. In contrast, our method deliberately avoids any use of demographic details to mitigate potential biases inherent in LLM training. Additionally, the reliance on demographic information may not always be feasible, which makes our approach more broadly applicable. Another key difference with \citet{hasan2024llm-gem}'s prompting strategy is the use of multiple input-output examples: while they rely on few-shot prompting with multiple example pairs to elicit LLM output in a consistent style, our approach does not require such examples, yet still achieves consistent outputs. Furthermore, they experimented solely on the GPT-3.5 LLM, whereas we explore both Llama 3 70B \citep{grattafiori2024llama3herdmodels} and GPT-4 \citep{openai2024gpt4technicalreport} LLMs in a recent dataset.

\subsection{Learning with Label Noise}\label{subsec:lit-noise}
Noise-robust learning has been extensively studied in classification settings, especially for computer vision, leaving textual regression tasks under-explored \citep{wang2022noisy}. Such learning algorithms were demonstrated either explicitly through dedicated methods \citep{englesson2024robust,natarajan2013learning,garg2021towards,li2022an} or implicitly as part of broader semi-supervised learning frameworks \citep{zhang2021flexmatch,sohn2020fixmatch}. Dedicated methods such as \citep{garg2021towards,li2022an} address noise by a de-noising loss function based on the usual cross-entropy loss function in classification. Semi-supervised methods \citep{zhang2021flexmatch,sohn2020fixmatch} generate pseudo-labels based on class probabilities produced by \texttt{sigmoid} or \texttt{softmax} activations. Both categories of methods have shown strong performance in classification tasks; however, they are not directly applicable to regression, where the target space is continuous and lacks discrete output logits. Our \textit{in-vitro} approach operates natively in the regression setting with the usual regression loss function and does not require any conversion to pseudo-class probabilities.

The study of label noise in textual regression remains limited. \citet{wang2022noisy}'s approach iteratively identifies noisy examples and applies one of three strategies: discarding noisy data points, substituting noisy labels with pseudo-labels, or resampling clean instances to balance the dataset. While effective in identifying extreme outliers, \citet{wang2022noisy} stated that their approach struggles in detecting mild disagreements. They further noted that their method performs worse in general-purpose datasets, compared to knowledge-dense domains, such as clinical notes and academic papers. Our approach leverages LLMs as an external ``teacher'' to directly correct noisy labels in a \emph{single} pass. Given the general-purpose nature of LLMs, our approach holds the potential to be effective across different domains.



\subsection{Empathy Computing}
Empathy computing is an emerging field, with significant advancements in textual empathy prediction \citep{hasan2024empathy}. For a detailed overview of its progress, we refer to a recent systematic literature review by \citet{hasan2024empathy}. 

In textual empathy computing, the most widely studied context is detecting people's empathy in response to newspaper articles. The Workshop on Computational Approaches to Subjectivity, Sentiment \& Social Media Analysis (WASSA) Shared Tasks (2021--2024) have spurred various approaches leveraging PLMs on this task. Most approaches predominantly employed fine-tuning PLMs, with RoBERTa being the most preferred PLM \citep{giorgi2024findings,qian2022surrey,lahnala2022caisa,chen2022iucl,plaza2022empathy,wang2023ynu,barriere2023findings,gruschka2023caisa,vasava2022transformer,kulkarni2021pvg,srinivas2023team,lu2023hit,frick2024fraunhofer}. Some studies have explored other BERT-based PLMs \citep{ghosh2022team,butala2021team,hasan2023curtin,numanoglu2024empathify} or ensemble strategies combining multiple PLMs \citep{mundra2021wassa,lin2023ncuee,chavan2023pict}. The suitability of fine-tuning RoBERTa is further validated by \citet{qian2022surrey}, who reported that simple fine-tuning of RoBERTa outperformed more complex multi-task learning in textual empathy computing. Overall, fine-tuning PLMs has emerged as the predominant approach for this task, with RoBERTa being the leading model \citep{hasan2024empathy}.

More recently, LLMs have been explored for textual empathy prediction through rephrasing text for data augmentation \citep{lu2023hit,hasan2024llm-gem}, fine-tuning \citep{li2024chinchunmei} and prompt engineering \citep{kong2024ru}. \citet{hasan2024llm-gem} adds multi-layer perception layers on top of a RoBERTa PLM to process demographic data, while \citet{li2024chinchunmei}'s fine-tuning of LLM demands significant computational resources. Unlike these methods, our approach leverages LLM-generated labels to enhance fine-tuning of a standard RoBERTa PLM, without demographic data or high computational costs.

\section{Method}

\subsection{Problem Formulation}\label{subsec:problem}
Let $\mathcal{D} = \{(x_i, y_i)\}_{i=1}^N$ represent a dataset, where $x_i$ is the $i$-th input text sequence, and $y_i \in \mathbb{R}$ denotes corresponding continuous empathy score. Empathy, being a psychological construct, is challenging to annotate due to subjectivity. Consequently, the target variable $y_i$ often suffers from noise, which is particularly significant in crowdsourced annotations (refer to \Cref{subsubsec:res-noise-mitig} for evidence of noise). We denote the noisy ground-truth empathy score as $\tilde{y}_i$, which serves as a proxy for the true, unobserved empathy score $y_i$. Thus, the dataset can be reformulated as $\mathcal{D} = \{(x_i, \tilde{y}_i)\}_{i=1}^N$. Our goal is to develop a model $\mathcal{F}: \mathcal{X} \to \mathbb{R}$, where $\mathcal{X}$ is the space of text sequences, such that $\mathcal{F}(x_i)$ accurately estimates $y_i$. 

The dataset $\mathcal{D}$ is randomly partitioned into three non-overlapping subsets: a training set $\mathcal{D}_{\text{train}} = \{(x_i, \tilde{y}_i)\}_{i=1}^{N_{\text{train}}}$, used to train the models; a validation set $\mathcal{D}_{\text{val}} = \{(x_j, \tilde{y}_j)\}_{j=1}^{N_{\text{val}}}$, used for optimising the training and tuning hyperparameters; and a hold-out test set $\mathcal{D}_{\text{test}} = \{(x_k, \tilde{y}_k)\}_{k=1}^{N_{\text{test}}}$, reserved for final model evaluation. The reserved $D_{\text{test}}$ has not been altered in any way through the experiments.

Large language models (LLMs) can be leveraged to improve training in such noisy scenarios, specifically to assist a smaller pre-trained language model (PLM) in better estimating the empathy score. We consider two \textit{in-vitro} approaches for leveraging LLM-generated labels to enhance model training. While such LLM use incurs non-trivial computational or API costs, it is a one-time expense during data preprocessing. All subsequent training and inference are performed using smaller PLMs, which keeps the overall resource requirement comparable to standard textual regression workflows.

\subsection{Applications of Large Language Model in-Vitro}

\begin{figure*}[t!]
    \centering
    \includegraphics[width=0.95\textwidth,trim={0.7cm 0 2.5cm 0},clip]{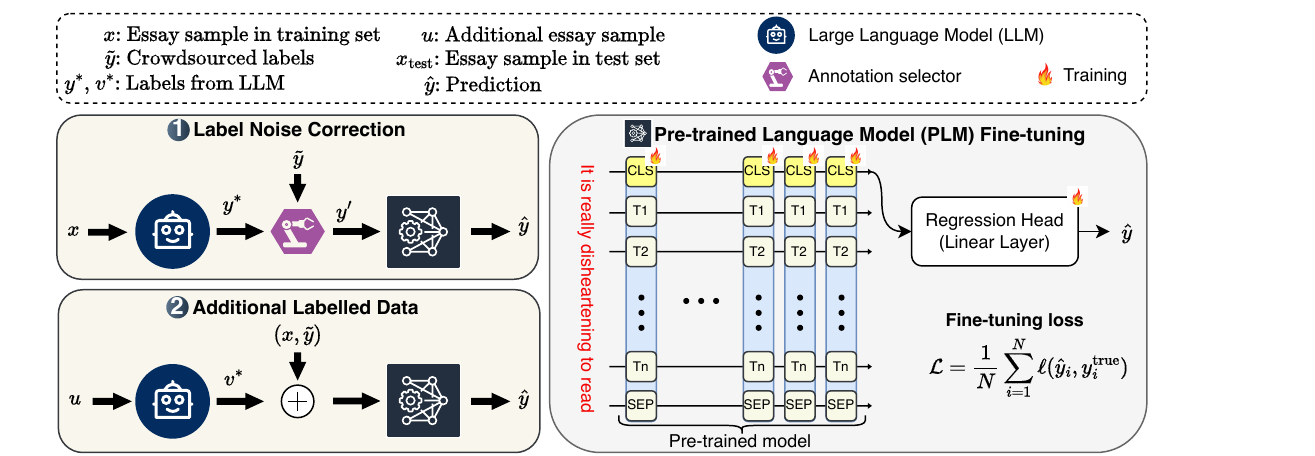}
    \caption{Overview of our proposed \textit{in-vitro} applications of large language models (LLMs) for enhancing textual empathy prediction with pre-trained language models (PLMs). Application 1 involves correcting noisy labels in an existing dataset using an LLM. Application 2 utilises an LLM to label additional text data, which is then added to the existing training dataset.}
    \label{fig:arch}
\end{figure*}

Our proposed framework leverages LLM for empathy prediction, as illustrated in \Cref{fig:arch}. The first application reduces label noise, while the second application increases the amount of training data by incorporating additional labelled data from LLM. Improved training data from these two applications is fed to a pre-trained language model (PLM) for final empathy prediction. 


\subsubsection{Prompt Design}\label{subsubsec:prompt}

\begin{table*}[t!]
\centering
\caption{Plain vs scale-aware prompt templates to generate labels from LLM.}
\label{tab:prompt}
\begin{tabularx}{\linewidth}{@{}
>{\raggedright\arraybackslash}p{0.8cm}
>{\raggedright\arraybackslash}X
>{\raggedright\arraybackslash}X
@{}}\toprule
\textbf{Scheme} & \textbf{System prompt} & \textbf{User prompt} \\ \midrule
Plain
& Your task is to measure the empathy of individuals based on their written essays.\newline
Human subjects wrote these essays after reading a newspaper article involving harm to individuals, groups of people, nature, etc. The essay is provided to you within triple backticks.
& Essay: \`{}\`{}\`{}\{\texttt{essay}\}\`{}\`{}\`{}\newline
Now, provide empathy score between 1.0 and 7.0, where a score of 1.0 means the lowest empathy, and a score of 7.0 means the highest empathy.\newline
You must not provide any other outputs apart from the scores \\ \midrule

Scale-aware
& Your task is to measure the empathy of individuals based on their written essays.\newline
You will assess empathy using Batson's definition, which specifically measures how the subject is feeling each of the following six emotions: sympathetic, moved, compassionate, tender, warm and softhearted.\newline
Human subjects wrote these essays after reading a newspaper article involving harm to individuals, groups of people, nature, etc. The essay is provided to you within triple backticks.
& Essay: \`{}\`{}\`{}\{\texttt{essay}\}\`{}\`{}\`{}\newline
Now, provide scores with respect to Batson's empathy scale. That is, provide scores between 1.0 and 7.0 for each of the following emotions: sympathetic, moved, compassionate, tender, warm and softhearted.\newline
You must provide comma-separated floating point scores, where a score of 1.0 means the individual is not feeling the emotion at all, and a score of 7.0 means the individual is extremely feeling the emotion.\newline
You must not provide any other outputs apart from the scores. \\ \bottomrule
\end{tabularx}
\end{table*}

To generate labels from LLM, one could instruct it to directly output the label (referred to as the \textit{plain} prompting scheme). Instead, we design a \textit{scale-aware} scheme, which includes subscales of the original labelling protocol. As presented in \Cref{tab:prompt}, each prompting scheme is structured into two primary components: the system prompt, which defines the task of the LLM and establishes the expected behaviour, and the user prompt, which contains specific input texts we want to annotate and the range of the outputs.

The motivation behind leveraging a scale-aware scheme includes natural alignment with the Psychology-grounded labelling protocol that was used for crowdsourced raters in the original study. For example, the NewsEmp datasets used Batson's Empathy scale, which has six subscales, and so our scale-aware prompt instructs the LLM to provide scores on the subscales (\Cref{tab:prompt}).
The difference between the human- and LLM-generated labelling is minimal, with the only difference being who labels the data. Therefore, following the same protocol designed for crowdsourced raters, LLM outputs across the subscales are averaged to calculate a single empathy score $y^*$. The following subsections present details about how we use these LLM labels in empathy computing.

\subsubsection{Application 1: Noise Mitigation in Labels}\label{subsubsec:noise-mitigation}
The LLM-generated labels $y^*$ are used to identify and replace potentially noisy samples in the original crowdsourced annotation $\tilde{y}$. Noisy samples are identified based on the difference between $\tilde{y}$ and $y^*$. Like \citep{hasan2024llm-gem}, a revised label $y'_i$ is defined as:
\begin{equation}
y'_i = 
    \begin{cases} 
    y_i^* & \text{if } |\tilde{y}_i - y_i^*| > \alpha \\
    \tilde{y}_i & \text{otherwise},
    \end{cases}
\end{equation}
where $\alpha$ is a predefined threshold, referred to as the \textit{annotation selection threshold}, which determines which label to use for which sample. 

This selection threshold can be any real number between 0 and the range of the empathy score (\ie $7-1=6$ for the NewsEmp dataset). A smaller $\alpha$ results in a more aggressive correction (replacing labels even for small differences). This could, however, lead to a larger distribution mismatch between the train and the test sets because the hold-out test set uses uncorrected crowdsourced labels. A model trained with a smaller $\alpha$ can, therefore, struggle to generalise on the hold-out test set.

Conversely, a larger $\alpha$ is a more conservative correction (replacing labels only at larger differences). Theoretically, a model trained with a larger $\alpha$ should generalise better on the test set because of a comparatively small distribution shift. This way, the model avoids training on crowdsourced labels that have a large deviation from LLM labels and, at the same time, enjoys the benefit of within-distribution crowdsourced labels that have slight deviations. 


The revised dataset $\mathcal{D}'_{\text{train}}  = \{(x_i, y'_i)\}$ is then used to train a PLM $\mathcal{F}_{y'}$. We hypothesise that the performance of $\mathcal{F}_{y'}$, trained on the mixture of $\tilde{y}$ and $y^*$, is better than $\mathcal{F}_{\tilde{y}}$, which is trained solely on $\tilde{y}$.

\subsubsection{Application 2: Additional Data Labelled by LLM}
Since deep learning models benefit from additional data, we propose to utilise LLM to get additional training data. While common LLM-based data augmentation techniques, such as paraphrasing \citep{vasava2022transformer,hasan2024llm-gem} and summarising \citep{hasan2023curtin}, are well-documented in the literature, our approach goes a step further. Specifically, we use an LLM to label new essays following the same annotation protocol as our target domain. This method, therefore, enables the integration of any similar data points into the training process.

Mathematically, we prompt LLM to annotate new text samples $u$ and make a new dataset $\mathcal{D}_\text{llm} = \{(u_i, v_i^*)\}_{i=1}^M$ with empathy scores $v^*$. These new data points could be any text similar to the essays $x$, but it may not have any prior empathy labels. We then annotate it in the same scale of $y$ using LLM. This additional data is combined with $\mathcal{D}_{\text{train}}$ to create an extended training set:
\begin{equation}
\mathcal{D}_{\text{train+}} = \mathcal{D}_{\text{train}} \cup \mathcal{D}_\text{llm} = \{(x_i, \tilde{y}_i)\}_{i=1}^{N_{\text{train}}} \cup \{(u_i, v_i^*)\}_{i=1}^M.
\end{equation}

Models trained on $\mathcal{D}_{\text{train+}}$ are expected to outperform models trained solely on $\mathcal{D}_{\text{train}} $ when evaluated on the hold-out test set $\mathcal{D}_{\text{test}}$. The extended dataset would enable the model to see a more diverse set of training examples, which should improve the model's ability to generalise on unseen test data.

Similar to labelling training data required for our proposed applications, LLMs can be prompted directly to generate empathy labels for the test set $D_\text{test}$. This zero-shot prediction leverages LLM's pre-trained knowledge without requiring further fine-tuning. Compared to the other two applications, zero-shot prediction relies heavily on the inherent capabilities of the LLM.


\subsection{Prediction using Pre-trained Language Model}
Fine-tuning a pre-trained language model (PLM) is a widely adopted approach in the empathy computing literature \citep{hasan2024empathy}. Accordingly, we utilise the dataset refined through LLM-based approaches to fine-tune a PLM. Each text sequence $x_i$ is first encoded into a contextual representation that serves as an aggregate sequence representation:
\begin{equation}
h_i^{\text{[CLS]}} = \text{PLM}(x_i; \theta),
\end{equation}
where $\theta$ are the parameters of the PLM, and $h_i^{\text{[CLS]}} \in \mathbb{R}^d$ is the [CLS] token representation. This pooled [CLS] representation is then passed through a linear regression head to predict the continuous empathy score:
\begin{equation}
\hat{y}_i = \mathcal{F}(h_i^{\text{[CLS]}}; \phi) = W h_i^{\text{[CLS]}} + b,
\end{equation}
where $\phi = {W \in \mathbb{R}^{1 \times d}, b \in \mathbb{R}}$ denotes the learnable parameters of the linear layer. The model is trained to minimise the discrepancy between predicted scores $\hat{y}_i$ and target scores $y_i^\text{true}$ across the dataset:
\begin{equation}
\mathcal{L} = \frac{1}{N} \sum_{i=1}^N \ell(\hat{y}_i, y_i^\text{true}),
\end{equation}
where $\ell(\cdot, \cdot)$ is the mean squared error (MSE) loss function, and $N$ is the number of training examples. The ground truth $y_i^\text{true}$ refers to the mixed labels $y'$ for Application 1, while for Application 2, it refers to crowdsourced labels $y$ for existing dataset $\mathcal{D}_\text{train}$ or LLM-provided labels $v^*$ for additional data $\mathcal{D}_\text{llm}$. The evaluation is always conducted on the original held-out dataset $\mathcal{D}_{\text{test}}$.

 

\begin{algorithm}[!t]
    \caption{Leveraging LLM in Empathy Detection}
    \label{alg:empathy_llm}
    \begin{algorithmic}[1]
        \Require Dataset $\mathcal{D} = \{(x_i, \tilde{y}_i)\}_{i=1}^N$, annotation selection threshold $\alpha$, additional unlabelled data $\mathcal{U} = \{u_i\}_{i=1}^M$
        \Ensure Empathy predictions $\hat{y}$

        \State \textbf{Partition} $\mathcal{D}$ into $\mathcal{D}_{\text{train}}$, $\mathcal{D}_{\text{val}}$ and $\mathcal{D}_{\text{test}}$

        \If{Application 1} 
            \State \Goto{app1}
        \ElsIf{Application 2}
            \State \Goto{app2}
        \EndIf

        \LComment{Application 1: label noise correction}\label{app1}
        \For{each $i$ in $\mathcal{D}_{\text{train}}$}
            \State Query LLM to generate label $y_i^*$ using scale-aware prompt
            \State $
                \text{Update label: } y'_i \gets \begin{cases} 
                    y_i^*, & \text{if } |\tilde{y}_i - y_i^*| > \alpha \\
                    \tilde{y}_i, & \text{otherwise}
                \end{cases}
            $
        \EndFor
        \State Form revised dataset $\mathcal{D}'_{\text{train}}  = \{(x_i, y'_i)\}_{i=1}^{N_{\text{train}}}$
        \State \Goto{plm}

        \LComment{Application 2: additional data labelled by LLM}\label{app2}
        \For{each $u_i$ in $\mathcal{U}$}
            \State Query LLM to generate label $v_i^*$ for $u_i$ using the same prompt
        \EndFor
        \State Form additional dataset $\mathcal{D}_{\text{llm}} = \{(u_i, v_i^*)\}_{i=1}^M$
        \State Combine datasets: $\mathcal{D}_{\text{train+}} = \mathcal{D}_{\text{train}} \cup \mathcal{D}_{\text{llm}}$
        \State \Goto{plm}

        \LComment{Prediction using pre-trained language model (PLM)}\label{plm}
        \State Fine-tune PLM $\mathcal{F}_{\theta}$ on $\mathcal{D}'_{\text{train}}$ or $\mathcal{D}_{\text{train+}}$:
        \begin{equation*}
            \hat{y}_i = \mathcal{F}_{\theta}(x_i) = W h_i^{\text{[CLS]}} + b
        \end{equation*}
        \State Evaluate $\mathcal{F}_{\theta}$ on $\mathcal{D}_{\text{test}}$

        \State \Return final predictions $\hat{y}$
    \end{algorithmic}
\end{algorithm}

\Cref{alg:empathy_llm} presents the overall workflow of our proposed approaches in empathy detection. After partitioning the dataset into training, validation and test subsets, one can choose between Application 1 and 2, as they are mutually exclusive. For Application 1 (label noise correction), the LLM is queried with scale-aware prompts to generate refined labels. If the difference between the original label and the LLM-generated label exceeds a threshold, the label is updated; otherwise, the original label is retained. The revised dataset is then used for PLM fine-tuning. For Application 2 (leveraging additional unlabelled data), the LLM is queried to generate labels for the unlabelled data, which is then combined with the training set to form an extended dataset. In both cases, a PLM is fine-tuned on the revised or extended dataset. The fine-tuning involves optimising the PLM to predict empathy scores based on input text embeddings, followed by evaluating its performance on the hold-out crowdsourced test set.

\section{Experiments and Results}

\subsection{Dataset and Associated Challenges}\label{subsec:data}
\citet{buechel2018modeling} marked an important step in understanding how individuals empathise with others or nature. They designed a crowdsourced approach where participants read newspaper articles depicting scenarios of harm to people or nature, and wrote about their emotional responses. The overarching aim was to capture individuals' reactions to adverse situations faced by others. This dataset, released in 2018, was the first of its kind, following which subsequent datasets were built. We refer to these datasets collectively as \textit{NewsEmp} series, as their central objective is to measure empathy elicited by newspaper articles.

The second NewsEmp dataset was released in 2022, in which \citet{tafreshi2021wassa} employed 564 subjects reading 418 news articles, which led to a total of 2,655 samples distributed into training, validation and test splits. Another significant change appeared in the NewsEmp\textsubscript{23} dataset \citep{omitaomu2022empathic}, which uses only the top 100 most negative articles from the pool of 418 news articles. Collectively, these datasets have become the most widely used dataset for benchmarking empathy detection approaches \citep{hasan2024empathy}. This popularity comes from their usage in the long-standing empathy detection challenge organised under the ``Workshop on Computational Approaches to Subjectivity, Sentiment \& Social Media Analysis (WASSA)'' series \citep{tafreshi2021wassa,barriere2022wassa,barriere2023findings,giorgi2024findings}. In particular, the WASSA 2021 \citep{tafreshi2021wassa} and WASSA 2022 \citep{barriere2022wassa} challenges utilised the NewsEmp\textsubscript{22} dataset, while WASSA 2023 \citep{barriere2023findings} and the WASSA 2024 \citep{giorgi2024findings} utilised the NewsEmp\textsubscript{23} and the latest NewsEmp\textsubscript{24} datasets, respectively. \Cref{tab:data-stat} presents the statistics of the three datasets used in this study.

\begin{table}[!t]
    \centering
    \caption{Statistics of the datasets used in this study.}
    \label{tab:data-stat}
    \begin{tabular}{l*6c} \toprule
        \textbf{Name} & \textbf{\# Train} & \textbf{\# Validation} & \textbf{\# Test}  & \textbf{\# Total} \\ \midrule
        NewsEmp\textsubscript{22} \citep{tafreshi2021wassa} & 1,860 & 270 & 525 & 2,655 \\
        NewsEmp\textsubscript{23} \citep{omitaomu2022empathic} & 792 & 208 & 100 & 1,100 \\
        NewsEmp\textsubscript{24} \citep{giorgi2024findings} & 1,000 & 63 & 83 & 1,146 \\ \bottomrule
    \end{tabular}
\end{table}

Perhaps due to the iterative nature of the datasets, there is overlap among some of these datasets. We found that the entire NewsEmp18 dataset is included in the training set of NewsEmp\textsubscript{22}, and the entire NewsEmp\textsubscript{23} training and validation sets appear in the NewsEmp\textsubscript{24} training set. In this study, we primarily use NewsEmp\textsubscript{22} and NewsEmp\textsubscript{24} datasets and partly NewsEmp\textsubscript{23} datasets. 

Although excluding NewsEmp\textsubscript{23} would have been feasible, prior research \citep{giorgi2024findings} achieved state-of-the-art results on the NewsEmp\textsubscript{24} dataset by combining NewsEmp\textsubscript{22}, NewsEmp\textsubscript{23} and NewsEmp\textsubscript{24} datasets to train their model. To ensure a fair comparison, we also report findings based on models trained using the combined three datasets. 

While this combination may seem unusual due to the overlap, it can be beneficial for improving predictions on the NewsEmp\textsubscript{24} test split. Very likely, this test split has a similar distribution to its own training set compared to another dataset's (NewsEmp\textsubscript{22}) training set, so including duplicated samples from the NewsEmp\textsubscript{24} training set allows the model to see more samples with a similar distribution.

It is worth noting that if we aim to evaluate a model on the NewsEmp\textsubscript{23} dataset, caution is necessary when combining datasets. One interesting finding on NewsEmp datasets is that -- although not explicitly stated in the studies \citep{omitaomu2022empathic,barriere2023findings,giorgi2024findings} reporting the datasets -- 44 out of 100 test samples in the NewsEmp\textsubscript{23} dataset are also present in the NewsEmp\textsubscript{24} validation set. Due to this data leakage, a model trained on the NewsEmp\textsubscript{24} validation set would, therefore, inflate performance on the NewsEmp\textsubscript{23} test split. To verify this, we trained a model using NewsEmp\textsubscript{24} training and validation sets, which gives a PCC of 0.576, outperforming the state-of-the-art PCC of 0.563 in NewsEmp\textsubscript{23} test split \citep{hasan2024llm-gem}. To prevent misleading results in future research, we highlight this overlap here and recommend exercising caution when combining datasets. Throughout our experiments, we ensure that there is no data leakage between training/validation and test splits.

We compare the performance of our proposed LLM-based approaches across various dataset combinations, including NewsEmp\textsubscript{24}, NewsEmp\textsubscript{23} and NewsEmp\textsubscript{22}. We then benchmark our work against the evaluation metrics reported by others on the NewsEmp\textsubscript{24} dataset. This dataset was chosen because it is the most recent in this series, and it includes the NewsEmp\textsubscript{23} dataset within it. Additionally, the ground truth for the NewsEmp\textsubscript{24} test split is publicly available, which is essential for calculating different metrics, while the ground truth for the other datasets is publicly unavailable.

\subsection{Evaluation Metric}\label{subsec:eval}
Pearson correlation coefficient (PCC) is the single metric used in the literature for evaluating empathy computing models across the NewsEmp datasets \citep{hasan2024empathy}. While it measures linear relationship between predicted and true values, it does not account for the \textit{magnitude} of errors, meaning predictions can have a perfect correlation with true values while being consistently offset (\eg predictions of 1, 2, and 3 corresponding to ground truths of 5, 6, and 7 yields a PCC of 1). This issue undermines its reliability for assessing model accuracy. 

While PCC has been the only metric used in empathy computing literature on NewsEmp datasets, studies on other datasets sometimes use different metrics. For example, \citet{barros2019omg}, detecting empathy in an audiovisual dataset, adopted the concordance correlation coefficient (CCC) as their primary metric. Previously mentioned shortcomings of PCC could be solved using CCC, as it calculates both the linear relationship and the magnitude of prediction errors. It ensures that predictions are not only aligned with the trend of true values but also close in magnitude, penalising large errors.

Root mean square error (RMSE) appears to be another choice of evaluation as it directly captures prediction error. Overall, PCC, CCC and RMSE measure three distinct qualities of performance: PCC measures linear relationship, RMSE measures the magnitude of errors, and CCC considers both linearity and error magnitude.

\subsection{Implementation Details}\label{subsec:impl}
We access Llama 3 (version: \texttt{llama3-70b-8192}) through Groq API and GPT-4 (version: \texttt{gpt-4o}) through OpenAI API (last accessed on 30 December 2024). The \textit{temperature} and \textit{top\_p} parameters of the APIs control the randomness of LLM outputs during token sampling. To ensure deterministic and consistent labels from LLMs, we set \textit{temperature} to $0$ and \textit{top\_p} to $0.01$.

As pre-trained language models (PLMs), we fine-tune RoBERTa (version: \texttt{roberta-base}) \citep{liu2019roberta}, which has 125.7M trainable parameters, and DeBERTa (version: \texttt{deberta-v3-base}) \citep{he2023deberta}, which has 184M trainable parameters, from Huggingface \citep{wolf2020huggingfaces}. We apply a delayed-start early-stopping strategy that starts monitoring validation CCC after five epochs and stops training if the score does not improve for two successive epochs. Early stopping based on PCC was ineffective in our experience due to higher fluctuations of the PCC score, whereas CCC performed better due to its smoother behaviour. Since most experiments converged within 20 epochs with early stopping, we set the maximum number of training epochs to 20. Deterministic behaviour was enforced using the PyTorch Lightning framework to ensure reproducibility. We save the model checkpoint corresponding to the last epoch of training. 

We adopt reported hyperparameters from the original work \citep{liu2019roberta} reporting RoBERTa. Following their reported approach to fine-tuning RoBERTa for downstream tasks, we only tuned the learning rate and batch size for our task. Values of the hyperparameters are reported in \Cref{tab:hyperparam}. While experimenting with the DeBERTa PLM, we use the same hyperparameters as used for RoBERTa.

\begin{table}[t!]
    \centering
    \caption{Hyperparameters for model training.}
    \label{tab:hyperparam}
    \resizebox{\columnwidth}{!}{%
    \begin{tabular}{@{}lc|lc@{}}\toprule
        \textbf{Pramerter} & \textbf{Value} & \textbf{Pramerter} & \textbf{Value} \\ \midrule
        Optimiser & AdamW & Learning rate scheduler & Linear \\
        Learning rate & 3e-5 & Warmup ratio & 0.06 \\
        AdamW ($\beta_1$, $\beta_2$) & (0.9, 0.98) & Batch size & 16 \\
        AdamW $\epsilon$ & 1e-6 & Maximum epochs & 20 \\
        Weight decay & 0.1 & Max sequence length & 512 \\
    \bottomrule
    \end{tabular}}
\end{table}

Following \citep{liu2019roberta}, we report median statistics over five different random initialisations (seeds: 0, 42, 100, 999 and 1234). Since prior works on empathy computing on these datasets reported a single peak score of their model, we also report the peak score from these five runs\footnote{We define \textit{peak} score as the best score (maximum PCC, maximum CCC or minimum RMSE) across five random runs within a \textit{single} experimental setup. Another related terminology used throughout this paper is the \textit{best} score, which refers to the best scores across \textit{different} experimental setups.}. All experiments are conducted in Python 3 running on a single AMD Instinct™ MI250X GPU (64 GB).

\subsection{Main Results}
This section presents the quantitative results of our proposed applications of LLM as a service (LLMaaS) in empathy prediction. Unless otherwise stated, results are reported primarily using the RoBERTa PLM, with DeBERTa results explicitly specified.

\begin{table*}[!t]
    \centering
    \caption{Examples of mislabelled crowdsourced annotation, deviating from Batson's definition of empathy. The first example shows an essay with empathic elements, but the participant's annotation indicates the lowest empathy. The second example has the highest empathy score despite the essay lacking empathic content. LLM labels appear accurate and consistent between Llama and GPT.}
    \label{tab:sample-anno}
    \resizebox{1\textwidth}{!}{%
    \begin{threeparttable}
    \begin{tabularx}{1.2\textwidth}{@{}
    >{\raggedright\arraybackslash}p{17.5cm}
    >{\centering\arraybackslash}X
    >{\centering\arraybackslash}X
    >{\centering\arraybackslash}X
    @{}} \toprule
        \textbf{Essay} & \textbf{Crowd} & \textbf{Llama} & \textbf{GPT} \\ \midrule
        ``After reading the article, my \emp{heart just breaks for the people} that are affected by this. Not only are innocent people being killed daily but also little children as well as babies. \emp{These children do not deserve this} and \emp{it's sad} because they have their whole lives ahead of them. \emp{I really hope} that war will end one day although it is looking unlikely.'' & 1.0 & 6.4 & 6.08 \\ \midrule
        ``I read the article on the China mining disaster. There were 33 miners trapped in the mine. Only two of them survived. Officials stated whoever was responsible would be punished. Smaller mines were shut down immediately until further notice. China has always been known for the deadliest mining.'' & 7.0 & 1.83 & 1.67 \\ \bottomrule
    \end{tabularx}%
    \begin{tablenotes}
        \item Empathic expressions are highlighted in \emp{blue}.
        \item Labels are in a continuous range from 1 to 7, where 1 and 7 refer to the lowest and highest empathy, respectively.
    \end{tablenotes}
    \end{threeparttable}
    }
\end{table*}

\subsubsection{Noise Mitigation}\label{subsubsec:res-noise-mitig}

We first show evidence of noise in the NewsEmp\textsubscript{24} dataset. \Cref{tab:sample-anno} illustrates a comparison between human participants' and LLMs' assessments of empathy on a scale of 1 (lowest empathy) to 7 (highest empathy) in two example essays. It demonstrates interesting disparities between crowdsourced and LLM evaluations -- for instance, in one essay expressing deep emotional concern for affected people and children, the human rater assigned a relatively low score of 1.0, while the Llama and GPT LLMs rated it much higher at 6.4 and 6.08, respectively. Conversely, a less empathic account of a mining disaster received the maximum possible empathy score (7.0) from human raters but a much lower 1.83 and 1.67 from the Llama and GPT LLMs, respectively. Interestingly, both LLMs, despite differences in size and provider, produce highly consistent annotations, which further underscores the potential inaccuracies of the crowdsourced annotation.

We evaluate our proposed LLMaaS application for noise mitigation in three dataset configurations: NewsEmp\textsubscript{24} alone, NewsEmp\textsubscript{24+22}, and NewsEmp\textsubscript{24+23+22}. While combining the datasets, we combine their training and validation splits of the additional dataset with the training split of the base dataset for training the model. For example, the NewsEmp\textsubscript{24+22} experimental setup uses the training split of NewsEmp\textsubscript{24} and training and validation splits of the NewsEmp\textsubscript{22} dataset to train the model. In all cases, the model training is optimised for the validation split of the NewsEmp\textsubscript{24} dataset and finally evaluated on the hold-out NewsEmp\textsubscript{24} test split.

\begin{table}[t!]
    \centering
    \caption{Results of our LLM-based noise mitigation approach, evaluated on the NewsEmp\textsubscript{24} test set.}
    \label{tab:noise_mitig}
    \begin{threeparttable}
    \begin{tabular}{l*4c}\toprule
        \textbf{Labels} & $\mathbf{\alpha}$ & \textbf{PCC} $\uparrow$ & \textbf{CCC} $\uparrow$ & \textbf{RMSE} $\downarrow$ \\ \toprule
\multicolumn{5}{l}{\texttt{\textbf{Data: NewsEmp\textsubscript{24}, PLM: RoBERTa}}} \\ \cmidrule{2-5}
        CS & -- & $0.331(0.378)$ & $0.307(0.329)$ & $1.656(\textbf{0.066})$ \\ \cmidrule{2-5}
        CS \& Llama & 3.5 & $\underline{0.453}(0.462)$ & $\textbf{0.435}(0.455)$ & $1.604(0.087)$ \\
            & 4.0 & $0.384(0.464)$ & $0.378(0.454)$ & $1.647(\underline{0.075})$ \\ 
            & 4.5 & $0.421(0.482)$ & $0.392(0.463)$ & $\underline{1.566}(0.098)$ \\ \cmidrule{2-5}
        CS \& GPT & 3.5 & $\textbf{0.473}(\underline{0.509})$ & $\underline{0.431}(\textbf{0.496})$ & $\textbf{1.558}(1.499)$ \\
            & 4.0 & $0.415(\textbf{0.519})$ & $0.398(\underline{0.482})$ & $1.601(1.461)$ \\
            & 4.5 & $0.370(0.422)$ & $0.325(0.400)$ & $1.646(1.532)$ \\ \midrule[1pt]
\multicolumn{5}{l}{\texttt{\textbf{Data: NewsEmp\textsubscript{24}, PLM: DeBERTa}}} \\ \cmidrule{2-5}
        CS & -- & $0.447(0.481)$ & $0.398(0.420)$ & $1.481(1.441)$ \\ \cmidrule{2-5}
        CS \& Llama & 3.5 & $0.502(0.516)$ & $0.450(0.511)$ & $1.531(1.445)$ \\
            & 4.0 & $\underline{0.536}(\underline{0.576})$ & $\underline{0.500}(0.518)$ & $\underline{1.413}(1.369)$ \\
            & 4.5 & $0.476(0.525)$ & $0.433(0.479)$ & $1.470(1.398)$ \\ \cmidrule{2-5}
        CS \& GPT & 3.5 & $0.529(0.568)$ & $0.489(\underline{0.536})$ & $1.419(1.393)$ \\
            & 4.0 & $\textbf{0.558}(\textbf{0.596})$ & $\textbf{0.533}(\textbf{0.571})$ & $\textbf{1.408}(\textbf{1.326})$ \\
            & 4.5 & $0.526(0.554)$ & $0.484(0.521)$ & $1.425(\underline{1.341})$ \\ \midrule[1pt]
\multicolumn{5}{l}{\texttt{\textbf{Data: NewsEmp\textsubscript{24+22}, PLM: RoBERTa}}} \\ \cmidrule{2-5}
        CS & -- & $0.536(0.597)$ & $0.461(0.505)$ & $1.356(\textbf{0.042})$ \\ \cmidrule{2-5}
        CS \& Llama & 0.0 & $0.483(0.504)$ & $0.445(0.480)$ & $1.655(1.628)$ \\
            & 0.5 & $0.488(0.503)$ & $0.445(0.481)$ & $1.646(1.617)$ \\
            & 1.0 & $0.479(0.491)$ & $0.432(0.451)$ & $1.756(1.664)$ \\
            & 1.5 & $0.474(0.498)$ & $0.434(0.448)$ & $1.694(1.583)$\\
            & 2.0 & $0.455(0.519)$ & $0.421(0.453)$ & $1.661(1.575)$ \\
            & 2.5 & $0.502(0.547)$ & $0.447(0.456)$ & $1.622(1.585)$ \\
            & 3.0 & $0.543(0.571)$ & $\underline{0.507}(0.512)$ & $1.461(1.435)$ \\
            & 3.5 & $\underline{0.558}(0.612)$ & $0.496(0.559)$ & $1.389(0.084)$ \\
            & 4.0 & $\textbf{0.589}(\textbf{0.627})$ & $\textbf{0.516}(\underline{0.563})$ & $\textbf{1.338}(\underline{0.054})$ \\ 
            & 4.5 & $0.551(\underline{0.620})$ & $0.478(\textbf{0.575})$ & $1.378(0.100)$ \\ 
            & 5.0 & $0.551(0.605)$ & $0.478(0.520)$ & $\underline{1.348}(1.265)$ \\
            & 5.5 & $0.542(0.604)$ & $0.464(0.516)$ & $1.352(1.265)$ \\
            & 6.0 & $0.536(0.597)$ & $0.461(0.505)$ & $1.356(1.274)$ \\ \midrule[1pt]
\multicolumn{5}{l}{\texttt{\textbf{Data: NewsEmp\textsubscript{24+23+22}, PLM: RoBERTa}}} \\ \cmidrule{2-5}
        CS & -- & $0.528(0.551)$ & $0.469(0.498)$ & $1.380(0.086)$ \\ \cmidrule{2-5}
        CS \& Llama & 3.5 & $\underline{0.556}(0.573)$ & $\underline{0.511}(\underline{0.552})$ & $1.381(\underline{0.029})$ \\
            & 4.0 & $\textbf{0.574}(\underline{0.582})$ & $\textbf{0.529}(0.548)$ & $\textbf{1.333}(\textbf{0.021})$ \\
            & 4.5 & $0.548(\textbf{0.648})$ & $0.479(\textbf{0.597})$ & $\underline{1.346}(0.092)$ \\
    \bottomrule
    \end{tabular}
    \begin{tablenotes}
        \item CS -- crowdsourced labels; $\alpha$ -- annotation selection threshold.
        \item Reported metrics are in \texttt{median(peak)} format, calculated from five random initialisations.
        \item \textbf{Boldface} and \underline{underline} texts indicate the best and the second-best scores, respectively.
    \end{tablenotes}
    \end{threeparttable}
\end{table}

As presented in \Cref{tab:noise_mitig}, our LLM-based noise mitigation approach demonstrates consistent performance improvements across all configurations. For the NewsEmp\textsubscript{24} dataset configuration, we report results with both Llama- and GPT-generated labels. While we choose RoBERTa as the primary PLM, we also report results with DeBERTa PLM in this setup, which shows even better performance improvement. Specifically, GPT labels at $\alpha = 4.0$ achieve the best median PCC (0.558), CCC (0.533) and RMSE (1.408) scores. Overall, the performance improvement between Llama and GPT labels is comparable, with each achieving the best results in certain metrics. Since Llama is an open-weight LLM and freely available, we proceed with Llama for the remaining experiments in this application scenario.

Including the NewsEmp\textsubscript{22} dataset enhances performance further, with $\alpha = 4.0$ yielding the best median PCC (0.589) and median CCC (0.516). When combined with NewsEmp\textsubscript{23}, the baseline metrics remain comparable, but $\alpha = 4.0$ again delivers the highest median PCC (0.574) and median CCC (0.529), with RMSE achieving its lowest value of 1.333. Considering peak scores instead of median statistics across five runs, our approach also outperforms the baseline model by achieving the peak PCC of 0.648, CCC of 0.597 and RMSE of 0.021. As illustrated earlier in \Cref{fig:significance}, the performance improvements are \emph{statistically} significant.

The value of the threshold $\alpha$ controls the proportion of LLM and crowdsourced labels. As demonstrated earlier, a smaller value of $\alpha$ means having a higher amount of LLM labels, which may hurt the model's performance in the test set. This hypothesis is verified in \Cref{tab:noise_mitig}'s results on varying $\alpha$ on the NewsEmp\textsubscript{24+22} scenario, which shows that $\alpha = 3.5 \sim 4.5$ provides the best performance across the three dataset configurations.

Our noise mitigation approach outperforms the baseline in terms of median PCCs, CCCs and RMSEs, as well as peak PCCs and CCCs across all four configurations. Out of 24 test cases\footnote{2 types (median, peak) $\times$ 3 metrics $\times$ 4 configurations in \Cref{tab:noise_mitig}.}, only two cases (RoBERTa on NewsEmp\textsubscript{24} and NewsEmp\textsubscript{24+22}), show a better peak RMSE achieved by the baseline. This discrepancy can be primarily attributed to how training was controlled: we applied early stopping based on CCC to mitigate overfitting. Since CCC and RMSE are not strongly correlated \citep{khorram2019jointly,han2017hard}, early stopping by CCC does not necessarily optimise RMSE. 

Recent works in affective computing, including emotion recognition \citep{atmaja2021evaluation} and empathy computing \citep{hasan2024empathy}, preferred correlation-based metrics over error-based metrics. Our findings align with this trend: although the baseline approach incidentally achieves a better peak RMSE in two isolated runs, our method demonstrates more consistent and robust performance across all metrics, including median RMSE across five runs, which reflects typical behaviour rather than outliers. In terms of the choice of the primary metric, PCC only captures linear correlation but not error magnitude, while RMSE only captures error magnitude but not correlation; therefore, our recommendation is to consider CCC as the primary metric, as it captures the best of both worlds -- both linear correlation and error magnitude.



\subsubsection{Additional Data Labelled by LLM}\label{subsubsec:res-additional}
The first application described above demonstrates that additional training data helps achieve better performance. However, we may not always have the flexibility of having extra data labelled by human participants. This application, therefore, explores whether additional data labelled by LLM could help. 

We evaluate this application in two configurations: either NewsEmp\textsubscript{24} or NewsEmp\textsubscript{22} as the base dataset. While evaluating the model on the NewsEmp\textsubscript{24} dataset, we consider the NewsEmp\textsubscript{22} dataset as additional data and vice versa. Since we have both crowdsourced and LLM-generated labels for the additional data, we compare whether the additional data is labelled by (1) human participants or (2) LLM.

\begin{table}[t!]
    \centering
    \caption{Effect of additional labelled data.}
    \label{tab:ext-data-llm}
    \resizebox{1\linewidth}{!}{%
    \begin{threeparttable}
    \begin{tabular}{@{}l*3c@{}}\toprule
        \textbf{Training data} & \textbf{PCC} $\uparrow$ & \textbf{CCC} $\uparrow$ & \textbf{RMSE} $\downarrow$ \\ \toprule
\multicolumn{4}{l}{\texttt{\textbf{PLM: RoBERTa}}} \\ \addlinespace[2pt]
        NewsEmp\textsubscript{24} & $0.331(0.378)$ & $0.307(0.329)$ & $1.656(\underline{0.066})$ \\
         + Crowd-labelled NewsEmp\textsubscript{22} & $0.485(\textbf{0.594})$ & $0.439(\underline{0.480})$ & $\textbf{1.417}(0.093)$ \\
         + Llama-labelled NewsEmp\textsubscript{22} & $\textbf{0.513}(\underline{0.571})$ & $\textbf{0.490}(\textbf{0.523})$ & $\underline{1.484}(\textbf{0.059})$ \\
         + GPT-labelled NewsEmp\textsubscript{22} & $\underline{0.495}(0.549)$ & $\underline{0.446}(0.455)$ & $1.581(1.514)$ \\ \midrule[1pt]
\multicolumn{4}{l}{\texttt{\textbf{PLM: DeBERTa}}} \\ \addlinespace[2pt]
        NewsEmp\textsubscript{24} & $0.447(0.481)$ & $0.398(0.420)$ & $1.481(1.441)$ \\
         + Crowd-labelled NewsEmp\textsubscript{22} & $\underline{0.564}(\textbf{0.638})$ & $0.478(\underline{0.566})$ & $\textbf{1.329}(\textbf{1.233})$ \\
         + Llama-labelled NewsEmp\textsubscript{22} & $\textbf{0.581}(\underline{0.601})$ & $\textbf{0.535}(\textbf{0.584})$ & $\underline{1.445}(\underline{1.366})$ \\
         + GPT-labelled NewsEmp\textsubscript{22} & $0.554(0.596)$ & $\underline{0.493}(0.534)$ & $1.468(1.391)$ \\ \midrule[1pt]
\multicolumn{4}{l}{\texttt{\textbf{PLM: RoBERTa}}} \\ \addlinespace[2pt]
        NewsEmp\textsubscript{22} & $0.459(0.477)$ & $0.363(0.411)$ & $1.776(\underline{0.046})$ \\
         + Crowd-labelled NewsEmp\textsubscript{24} & $\underline{0.467}(\underline{0.478})$ & $\underline{0.392}(\textbf{0.435})$ & $\underline{1.756}(0.062)$ \\
         + Llama-labelled NewsEmp\textsubscript{24} & $\textbf{0.496}(\textbf{0.519})$ & $\textbf{0.429}(\underline{0.434})$ & $\textbf{1.729}(\textbf{0.034})$ \\
        \bottomrule
    \end{tabular}
    \begin{tablenotes}
        \item All evaluations use test splits, except for NewsEm22, where CCC and RMSE are computed on the validation split due to unavailable test labels.
        \item Reported metrics are in \texttt{median(peak)} format, calculated across five random initialisations.
        \item \textbf{Boldface} and \underline{underline} texts indicate the best and the second-best scores, respectively.
    \end{tablenotes}
    \end{threeparttable}}
\end{table}

\Cref{tab:ext-data-llm} reports the performance in both settings. When trained a RoBERTa model with the NewsEmp\textsubscript{24} dataset alone, it achieved a median PCC of 0.331 and 0.307 CCC. Additional NewsEmp\textsubscript{22} dataset labelled by human participants boosted performance to 0.485 PCC, 0.439 CCC and 1.417 RMSE. The same NewsEmp\textsubscript{22} dataset -- labelled by Llama LLM -- makes the highest median PCC of 0.513 and CCC of 0.490 while maintaining a competitive RMSE of 1.484. For this NewsEmp\textsubscript{24} setup, we further report results with the DeBERTa PLM, which shows even better performance across different metrics. Like our earlier experiment (Application 1), the use of either Llama or GPT LLMs yields similar performance in this setup, and we proceed with Llama LLMs for the rest of the experiments.

Experiments using NewsEmp\textsubscript{22} as the base dataset exhibited a similar trend to those with NewsEmp\textsubscript{24} as the base dataset. The model trained with LLM-generated labels achieves the best PCC (0.496) and CCC (0.429), as well as the lowest RMSE (1.729). Notably, the human-labelled data shows improvement over the base dataset but falls short of the performance achieved with LLM-labelled data. Overall, LLM labels are as good as crowdsourced labels, and additional data, either crowdsourced or LLM-labelled, boosts the performance. 

Compared to our first application scenario (mixed labels to reduce label noise), performance improvement from baseline in this application is statistically more significant across all three evaluation metrics (\Cref{fig:significance}). In particular, this application scenario demonstrates the highest level of statistically significant improvements in terms of PCC and CCC. This is likely because, in this scenario, the labels of the base training data remain unchanged, which is presumably of a similar distribution to the hold-out test set. The additional data provides extra supervision, which helps achieve a better score. Results using additional data labelled by LLM are better than using human labelling in most cases (12 out of 18 test cases), likely because of the higher quality of labels from LLM.

\begin{figure}[t!]
    \centering
    \includegraphics[width=0.8\columnwidth]{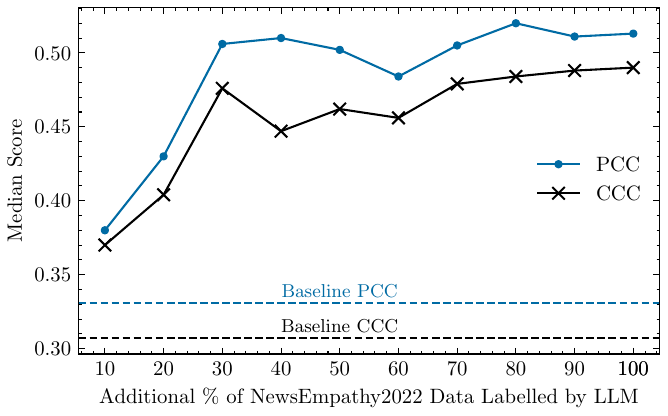}
    \caption{Median performance in the NewsEmp\textsubscript{24} test set with gradual increase of additional LLM-labelled data. \textit{Baseline} scores refer to the scores achieved using only NewsEmp\textsubscript{24} data.}
    \label{fig:llm_portion}
\end{figure}

To understand how the amount of additional data affects the performance, we gradually increase the amount of additional data and report model performance (\Cref{fig:llm_portion}). In each case, we randomly sample a percentage of the additional data ranging from 10\% to 100\%. Surprisingly, the performance increases most rapidly from 10 to 30\%, after which the improvement slows down.

\begin{figure*}[t!]
    \centering
    \includegraphics[width=0.95\textwidth]{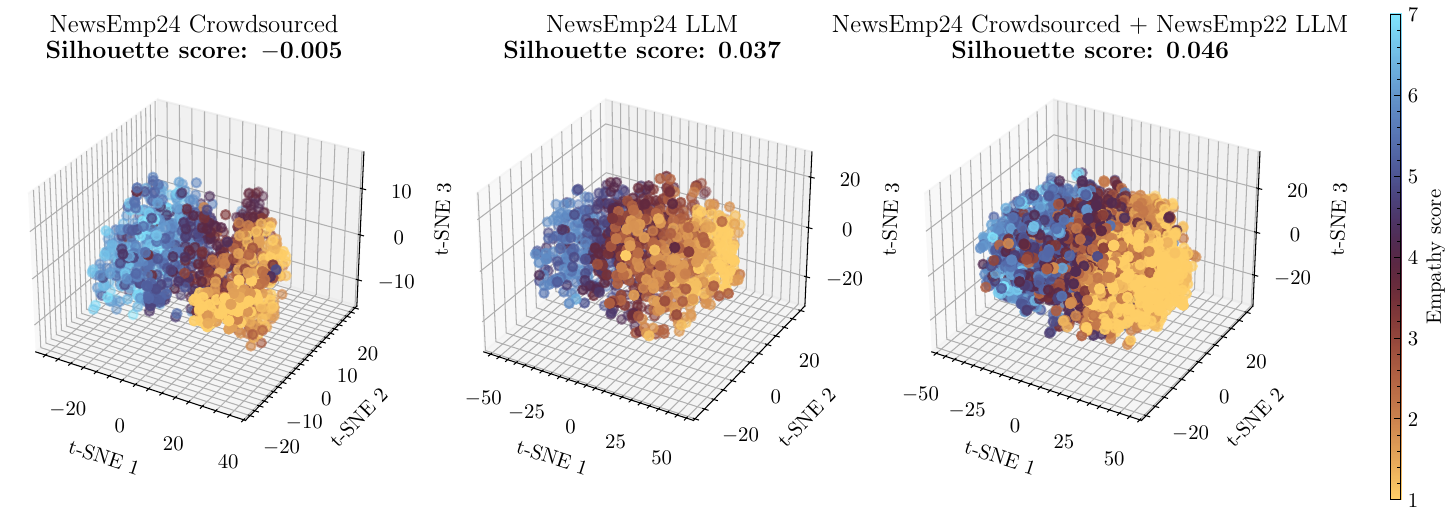}
    \caption{3D t-SNE visualisation and Silhouette scores on CLS embeddings from pre-trained language models fine-tuned using crowdsourced labels (\textbf{left}), LLM-generated labels (\textbf{middle}) and crowdsourced + additional LLM-labelled data (\textbf{right}). Continuous empathy labels are discretised into six bins to calculate Silhouette scores (higher is better), which suggests better clustering after integrating LLM-generated labels.
    }
    \label{fig:tsne}
\end{figure*}

\Cref{fig:tsne} presents 3D t-SNE visualisations of embeddings derived from different labelling schemes, alongside their clustering performance measured by the mean Silhouette score \citep{rousseeuw1987silhouettes} on the embeddings from PLM. Given the continuous nature of empathy labels in the datasets, we discretise them into six bins (1--2, 2--3, ..., 6--7) to calculate the metric. This score ranges from $-1$ to $+1$, with $-1$ being \textit{``misclassified''} and $+1$ being \textit{``well-clustered''} \citep{rousseeuw1987silhouettes}. 

Embeddings from crowdsourced labels exhibit a slightly dispersed distribution visually and the lowest Silhouette score  ($-0.005$), which suggests many samples are mislabelled (\ie label noise). In contrast, embeddings based on LLM labels display smoother distributions (especially the rightmost plot) and higher Silhouette scores ($0.037$ and $0.046$)\footnote{Note that although \Cref{fig:tsne} shows relative differences in Silhouette scores across embeddings, their absolute values remain low, presumably due to the discretisation of continuous labels (the clustered labels are not well-separated groups).}, which suggests many of the mislabelled samples are now correctly labelled (\ie reduction of label noise). The smoothest distribution and the highest Silhouette score on the NewsEmp\textsubscript{24} Crowdsourced + NewsEmp\textsubscript{22} LLM configuration can be attributed to extra supervision from the LLM-generated labels.

\subsection{Zero-Shot Prediction \& Demographic Biases}\label{subsec:zero-and-bias}
The most direct application of LLM is to predict empathy in a zero-shot manner, \ie without any training or fine-tuning of the LLM. \Cref{tab:zero-shot} reports the performances of zero-shot prediction across all validation and test splits.

\begin{table}[t!]
    \centering
    \caption{Zero-shot empathy prediction using LLMs.}
    \label{tab:zero-shot}
    \begin{threeparttable}
    \begin{tabular}{@{}*3l*3c@{}} \toprule
        \textbf{Dataset} & \textbf{LLM} & \textbf{Split} & \textbf{PCC $\uparrow$} & \textbf{CCC $\uparrow$} & \textbf{RMSE $\downarrow$} \\ \midrule
        NewsEmp\textsubscript{24} & Llama & Test & 0.441 & 0.436 & 1.731 \\
            &   & Validation & 0.502 & 0.457 & 1.952 \\ \cmidrule{2-6}
            & Llama\textsubscript{Plain} & Test & 0.405 & 0.358 & 1.859 \\ \cmidrule{2-6}
            & GPT & Test & 0.581 & 0.489 & 1.715 \\
            &   & Validation & 0.480 & 0.375 & 2.038 \\ \midrule
        NewsEmp\textsubscript{23} & Llama & Test & 0.380 & -- & -- \\
            &   & Validation & 0.108 & 0.108 & 2.18 \\ \midrule
        NewsEmp\textsubscript{22} & Llama & Test & 0.517 & -- & -- \\
            &   & Validation & 0.579 & 0.573 & 1.728 \\
    \bottomrule
    \end{tabular}
    \begin{tablenotes}
        \item Llama\textsubscript{Plain}: Llama with the plain prompt. All other entries use the scale-aware prompt.
        \item Ground truth of the NewsEm22 and NewsEmp\textsubscript{23} test splits are unavailable to calculate CCC and RMSE.
    \end{tablenotes}
    \end{threeparttable}
\end{table}

On the NewsEmp\textsubscript{24} test split, we compare plain prompting with our proposed scale-aware prompting scheme. The improved performance with the scale-aware prompt (PCC: $0.405 \rightarrow 0.441$; CCC: $0.358 \rightarrow 0.436$; RMSE: $1.859 \rightarrow 1.731$) demonstrates its effectiveness. Accordingly, we adopt the scale-aware prompt as the default in all our experiments.

\begin{figure*}[!th]
    \centering
    \includegraphics[width=0.95\linewidth]{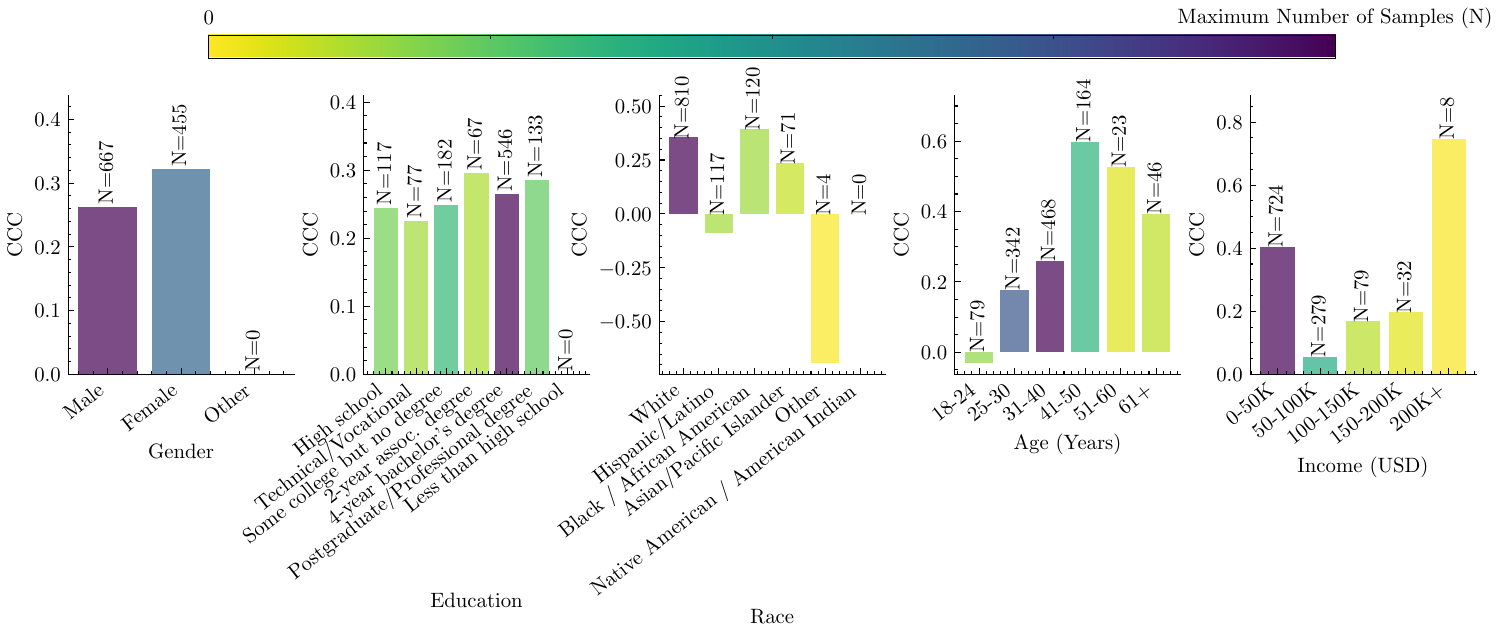}
    \caption{Number of samples and zero-shot (Llama) prediction performance across different demographic groups in the NewsEmp\textsubscript{24} dataset. CCC varies rapidly across different racial groups.}
    \label{fig:demog-distr}
\end{figure*}

We further examine how the agreement between crowdsourced annotation and LLM annotation varies across different demographic groups. For this analysis, we combine training, validation and test splits of the NewsEmp\textsubscript{24} dataset and compare crowdsourced and Llama-generated labels. As demonstrated in \Cref{fig:demog-distr}, both have similar levels of CCC in gender and education demographics. However, CCC changed wildly across race, age and income groups. In particular, it went to negatives in two race groups: ``Hispanic/Latino'' and ``Other'' categories, noting that there are only four samples in the ``Other'' category of race.

In terms of the number of samples across different demographics, we see that certain demographic groups (\eg ``4-year bachelor’s degree'' education, ``White'' race, and ``31-40'' age groups) are highly represented compared to their counterparts. Some demographics, for example, ``Less than high school'' education and ``Native American / American Indian'' are not represented in the dataset at all. Such imbalances can presumably introduce bias in the empathy detection model built from such biased datasets.

LLMs designed for empathy detection may exhibit biases across different demographic groups \citep{gabriel-etal-2024-ai}. A proper assessment of such bias would require accurate and unbiased ground truth labels to compare with. Note that \Cref{fig:demog-distr} compares LLM outputs against potentially noisy crowdsourced labels and so may not offer an objective assessment of bias. Instead, it reflects potential \textit{sources} of bias that may influence an empathy detection model, insofar as biased training data can propagate bias into the model's predictions. 

Our prompt to the LLM is \textit{demographic-unaware}, meaning it does not include any explicit demographic information. Theoretically, LLM outputs through such an unaware prompt should be less biased (because it is a blind evaluation) compared to a demographic-aware prompt \citep{gabriel-etal-2024-ai}. \citet{gabriel-etal-2024-ai} argued that a demographic-aware prompt may also result in less biased assessment, since the LLMs are being alerted to potential bias. They found mixed outcomes on mitigating bias through demographic-aware and unaware prompts across different LLMs \citep{gabriel-etal-2024-ai}. Mitigating bias is critical for real-world deployment, and so future exploration towards a universally applicable prompting strategy is warranted to mitigate demographic biases.




\subsection{Comparison with the Literature}
A quantitative comparison between our proposed framework and other empathy detection works in the literature on the NewsEmp\textsubscript{24} dataset is presented on \Cref{tab:comp}. The best performance in the literature is 0.629 PCC \citep{giorgi2024findings}, while our best performance is 0.648 PCC. Both \citet{giorgi2024findings} and we use the same amount of training data -- combined NewsEmp\textsubscript{24+23+22}.

\begin{table}[t!]
    \centering
    \caption{Comparison of our proposed model with the literature on the NewsEmp\textsubscript{24} test dataset.}
    \label{tab:comp}
    \begin{threeparttable}
    \begin{tabular}{ll*3c} \toprule
        \textbf{Approach} & \textbf{Base Model (Ref.)} & \textbf{PCC $\uparrow$} & \textbf{CCC $\uparrow$} & \textbf{RMSE $\downarrow$} \\ \midrule
        Training & BERT \citep{numanoglu2024empathify} & 0.290 & -- & -- \\
         & MLP \citep{chevi2024daisy} & 0.345 & -- & -- \\ 
         & RoBERTa \citep{frick2024fraunhofer} & 0.375 & -- & -- \\
         & \textit{Not mentioned} \citep{pereira-etal-2024-context} & 0.390 & -- & -- \\
         & Llama 3 8B \citep{li2024chinchunmei} & 0.474 & -- & -- \\
        
         & RoBERTa \citep{giorgi2024findings}& 0.629 & -- & -- \\
         & RoBERTa \citep{giorgi2024findings}\tnote{a} & 0.607 & 0.498 & \textbf{0.075} \\
         & RoBERTa (\textbf{Ours}) & \textbf{0.648} & \textbf{0.597} & 0.092 \\ \midrule
        Zero-shot & GPT 3.5 \citep{kong2024ru} & 0.523 & -- & -- \\
         & GPT 4  (\textbf{Ours}) & 0.581 & 0.489 & 1.715 \\
        \bottomrule
    \end{tabular}
    \begin{tablenotes}
        \item[a] Our implementation of the earlier SOTA work \citep{giorgi2024findings}.
    \end{tablenotes}
    \end{threeparttable}
\end{table}

The reported results in the literature are in terms of a single evaluation metric, which suggests the peak performance of their model. To compare in terms of other evaluation metrics in a similar setting of ours (five random initialisations), we implemented the state-of-the-art work \citep{giorgi2024findings}. The mismatch between our implementation and \citet{giorgi2024findings}'s reported result (0.607 vs 0.629) is likely due to hyperparameter choice. Having no public implementation of \citet{giorgi2024findings}, we chose default hyperparameters, apart from the minimal amount of hyperparameter details reported in their work. Our approach outperforms \citet{giorgi2024findings}'s results in terms of both PCC and CCC (\Cref{tab:comp}).

\subsection{Consistency and Inter-Rater Reliability}
LLMs are known to produce varying outputs across different API calls \citep{Ouyang_2024}. This variability could raise concerns about using LLM to label data as well as evaluating on the test set. We calculate two types of consistency: \textit{intra-LLM} consistency, which evaluates whether annotations generated by the same LLM model remain consistent across multiple API calls, and \textit{inter-LLM} consistency, which assesses whether annotations are consistent between two different LLMs.

\begin{table}[t!]
    \centering
    \begin{threeparttable}
    \caption{Consistency and inter-rater reliability among Llama, GPT and crowdsourced annotations on the NewsEmp\textsubscript{24} training set. The low reliability between LLMs and crowdsourced annotations, contrasted with the high reliability between two different LLMs, may suggest that the crowdsourced annotations are noisy.}
    \label{tab:reliab}
    \begin{tabular}{*2l*2c} \toprule
        \textbf{Annotator 1} & \textbf{Annotator 2} & \textbf{Krippendorff's Alpha} & \textbf{MAE $\pm$ SD} \\ \midrule
        Llama & Llama & 0.99 & $0.10\pm0.21$ \\
        Llama & GPT & 0.80 & $0.78\pm0.70$ \\
        Llama & Crowd & 0.27 & $1.72\pm1.34$ \\
        GPT & Crowd & 0.19 & $1.81\pm1.27$ \\
        \bottomrule
    \end{tabular}
    \begin{tablenotes}
        \item MAE -- mean absolute error; SD -- standard deviation of absolute error
    \end{tablenotes}
    \end{threeparttable}
\end{table}

To assess intra-LLM consistency, we label the NewsEmp\textsubscript{24} training set (1,000 samples) twice in the Llama LLM, using separate independent API calls. The results (\Cref{tab:reliab}) demonstrate \emph{``almost perfect''} agreement between the two annotation rounds, with a Krippendorff's Alpha (K-Alpha) \citep{krippendorff2019reliability} score of 0.99. Between GPT and LLM annotations (inter-LLM), a K-Alpha of 0.80 is achieved, which lies on the boundary between \emph{``substantial''} and \emph{``almost perfect''} reliability \citep{krippendorff2019reliability}. Such a high level of consistency, including inter-LLM consistency, suggests the effectiveness of our prompting strategy, which clearly specifies the expectations from the LLM.

As presented in \Cref{tab:reliab}, the inter-rater reliability between LLMs and crowdsourced annotations is notably lower than the reliability observed between LLMs. It potentially supports our hypothesis that crowdsourced annotations are inherently noisy.


\subsection{Human Preference Study: Crowdsourced vs LLM Labels}\label{subsec:human}
To evaluate the credibility of LLM-generated empathy scores relative to crowdsourced annotations, we conducted a small-scale human assessment study involving three co-authors (all PhD holders, including two mid-career and one senior academic) as independent assessors. The corresponding author designed the experiment, while the assessors were blind to the origin of each label (LLM or crowdsourced). We selected 20 samples that exhibited the largest disagreement between LLM and crowdsourced annotations to better test discernibility. Each assessor was shown an essay along with two empathy scores (randomised in order) and asked to choose the score that best reflected the empathy expressed in the essay, following Batson’s definition of empathy.

Across the 20 samples, 8 were unanimously rated in favour of the LLM-generated labels by all three assessors. In 10 other samples, two out of three assessors preferred the LLM labels. Only in 2 cases did the majority (2 of 3) select the crowdsourced label. Based on majority voting, LLM-generated labels were preferred in 18 out of 20 instances, suggesting that, at least in these high-discrepancy cases, LLM annotations aligned more closely with human expert judgment than the original crowdsourced labels.

\subsection{Limitations and Future Work}\label{subsec:limit}
As we note the inherent biases in LLM, it is crucial to exercise caution when using LLM-generated labels if such a system is to be deployed in real life. Zero-shot predictions are likely to exhibit greater bias, as LLMs may inherit biases from their training data. Therefore, downstream models should be trained on diverse and representative datasets that reflect the demographics in which empathy would be detected.

As this paper investigates how LLM-generated labels can help measure empathy by smaller PLMs, we restrict the LLMs to generating labels only. However, incorporating the reasoning behind these labels could enhance explainability and potentially further improve label quality. Future work may explicitly prompt LLMs to demand justification, employ chain-of-thought prompting to elicit reasoning \citep{wei2022chain}, or leverage LLMs having implicit reasoning capability (\eg OpenAI o3). Lastly, as all experiments are conducted on English datasets (NewsEmp series), exploring the multilingual adaptability of our methods remains an important direction for future work.

\section{Conclusion}
This work demonstrates the potential of large language models (LLMs) in addressing challenges in empathy computing through two \textit{in-vitro} applications: label noise reduction and training data expansion. Both applications resulted in statistically significant performance gains over baseline methods. The proposed framework outperformed state-of-the-art methods, achieving new benchmarks on a public empathy dataset with a Pearson correlation coefficient (PCC) of 0.648, among other metrics. Beyond the empirical results, this paper contributes a critical rethinking of evaluation practices in empathy computing, advocating for the adoption of the concordance correlation coefficient (CCC). The novel scale-aware prompting technique introduced here ensures alignment between LLM annotations and theoretical annotation protocols. We further highlight biases in the dataset across different demographic groups. Similar to the empathy detection dataset addressed in this paper, many other tasks, such as detecting depression, anxiety and mental health conditions, rely on questionnaire-based self-annotations. The proposed approach, therefore, opens exciting avenues for leveraging LLMs as complementary tools to enhance model training across different domains.

\section*{Acknowledgement}
This work was supported by resources provided by the Pawsey Supercomputing Research Centre with funding from the Australian Government and the Government of Western Australia.

\printbibliography



 





\end{document}